\documentclass[letterpaper, 10 pt, conference]{ieeeconf} 

\IEEEoverridecommandlockouts                        

\usepackage{graphics}
\usepackage{amsmath} 
\usepackage{amssymb}
\usepackage{cite}
\usepackage{tikz}
\usepackage{amsmath}
\usepackage{tabularx}
\usepackage{array}
\usepackage{booktabs}
\usepackage{bm}
\usepackage{algorithm}
\usepackage{algorithmic}
\usepackage{bbm}
\usepackage{hyperref}
\usepackage{units}
\usepackage{subfigure}
\usepackage{caption}

\usepackage{tablefootnote}

\definecolor{green}{RGB}{3,200,15}

\title{\LARGE \bf
Autonomous Ground Navigation in Highly Constrained Spaces: \\Lessons learned from The BARN Challenge at ICRA 2022
}
\author{\textbf{Competition Organizers}: Xuesu Xiao$^{1, 2, 3}$, Zifan Xu$^{3}$, Zizhao Wang$^{3}$, Yunlong Song$^{4}$,\\ Garrett Warnell$^{3, 5}$, Peter Stone$^{3, 6}$, Tingnan Zhang$^{7}$, 
\\\textbf{Finals Participants}: Shravan Ravi$^{3}$, Gary Wang$^{3}$, Haresh Karnan$^{3}$, Joydeep Biswas$^{3}$,\\  Nicholas Mohammad$^{8}$, Lauren Bramblett$^{8}$, Rahul Peddi$^{8}$, Nicola Bezzo$^{8}$, \\
Zhanteng Xie$^{9}$, and Philip Dames$^{9}$
\thanks{$^{1}$George Mason University
$^{2}$Everyday Robots
$^{3}$The University of Texas at Austin
$^{4}$University of Zurich
$^{5}$Army Research Laboratory
$^{6}$Sony AI 
$^{7}$Robotics@Google
$^{8}$University of Virginia
$^{9}$Temple University
}
}
\begin{document}

\maketitle
\thispagestyle{empty}
\pagestyle{empty}

\begin{abstract}

The BARN (Benchmark Autonomous Robot Navigation) Challenge took place at the 2022 IEEE International Conference on Robotics and Automation (ICRA 2022) in Philadelphia, PA. 
The aim of the challenge was to evaluate state-of-the-art autonomous ground navigation systems for moving robots through highly constrained environments in a safe and efficient manner.
Specifically, the task was to navigate a standardized, differential-drive ground robot from a predefined start location to a goal location as quickly as possible without colliding with any obstacles, both in simulation and in the real world. 
Five teams from all over the world participated in the qualifying simulation competition, three of which were invited to compete with each other at a set of physical obstacle courses at the conference center in Philadelphia. 
The competition results suggest that autonomous ground navigation in highly constrained spaces, despite seeming ostensibly simple even for experienced roboticists, is actually far from being a solved problem. 
In this article, we discuss the challenge, the approaches used by the top three winning teams, and lessons learned to direct future research. 

\end{abstract}
\section{The BARN Challenge Overview}
\label{sec::challenge}

Designing autonomous robot navigation systems has been a topic of interest to the robotics community for decades~\cite{rosmann2012trajectory, rosmann2013efficient, rosmann2015planning, quinlan1993elastic, fox1997dynamic}.
Indeed, there currently exist many such systems that allow robots to move from one point to another in a collision-free manner (e.g., open-source implementations in the Robot Operating System (ROS)~\cite{quinlan1993elastic, fox1997dynamic, rosmann2017integrated} with extensions to different vehicle types~\cite{rosmann2017kinodynamic}), which may create the perception that autonomous ground navigation is a solved problem.
This perception may be reinforced by the fact that many mobile robot researchers have moved on to orthogonal navigation problems~\cite{xiao2022motion} beyond the traditional metric (geometric) formulation that only focuses on path optimality and obstacle avoidance.
These orthogonal problems include, among others, learning navigation systems in a data-driven manner~\cite{pfeiffer2017perception, chiang2019learning, voilaharesh, optimfkd}, navigating in off-road~\cite{pan2020imitation, kahn2021badgr, karnan2022vi} and social contexts~\cite{mirsky2021prevention, mavrogiannis2021core, scand}, and multi-robot navigation~\cite{long2018towards, chen2017decentralized}.

However, autonomous mobile robots still struggle in many \emph{ostensibly} simple scenarios, especially during real-world deployment~\cite{irobot, everydayrobots, scout, starship}.
For example, even when the problem is simply formulated as traditional metric navigation so that the only requirement is to avoid obstacles on the way to the goal, robots still often get stuck or collide with obstacles when trying to navigate in naturally cluttered daily households~\cite{irobot}, in constrained outdoor structures including narrow walkways and ramps~\cite{scout, starship}, and in congested social spaces like classrooms, offices, and cafeterias~\cite{everydayrobots}.
In such scenarios, extensive engineering effort is typically required in order to deploy existing approaches, and this requirement presents a challenge for large-scale, unsupervised, real-world robot deployment.
Overcoming this challenge requires systems that can both successfully and efficiently navigate a wide variety of environments with confidence.

The Benchmark Autonomous Robot Navigation (BARN) Challenge~\cite{the_barn_challenge} was a competition at the 2022 IEEE International Conference on Robotics and Automation (ICRA 2022) in Philadelphia, PA that aimed to evaluate the capability of state-of-the-art navigation systems to solve the above-mentioned challenge, especially in highly-constrained environments, where robots need to squeeze between obstacles to navigate to the goal.
To compete in The BARN Challenge, each participating team needed to develop an entire software stack for navigation for a standardized and provided mobile robot.
In particular, the competition provided a Clearpath Jackal~\cite{clearpath_jackal} with a 2D 270\textdegree-field-of-view Hokuyo LiDAR for perception and a differential drive system with 2m/s maximum speed for actuation.
The aim of each team was to develop navigation software stack needed to autonomously drive the robot from a given starting location through a dense obstacle filed and to a given goal, and to accomplish this task without any collisions with obstacles or any human interventions.
The team whose system could best accomplish this task within the least amount of time would win the competition.
The BARN Challenge had two phases: a qualifying phase evaluated in simulation, and a final phase evaluated in a set of physical obstacle courses.
The qualifying phase took place before the ICRA 2022 conference using the BARN dataset~\cite{perille2020benchmarking}, which is composed of 300 obstacle courses in Gazebo simulation randomly generated by cellular automata.
The top three teams from the simulation phase were then invited to compete in three different physical obstacle courses set up by the organizers at ICRA 2022 in the Philadelphia Convention Center.

In this article, we report on the simulation qualifier and physical finals of The BARN Challenge at ICRA 2022, present the approaches used by the top three teams, and discuss lessons learned from the challenge that point out future research directions to solve the problem of autonomous ground navigation in highly constrained spaces.
\section{Simulation Qualifier}
\label{sec::simulation}
\begin{figure*}[t]
    \centering
    \includegraphics[width=2\columnwidth]{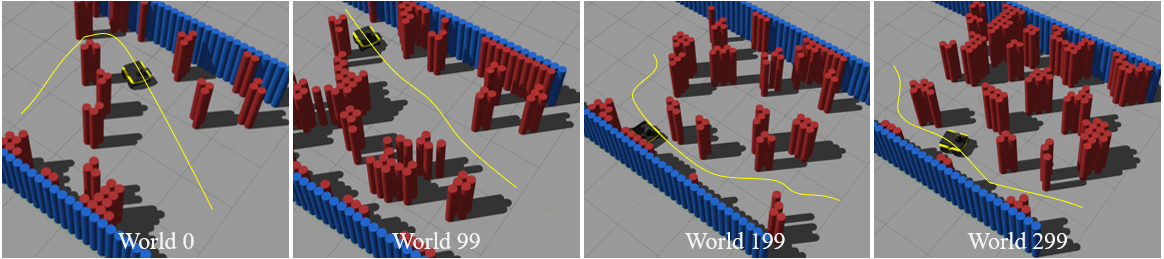}
    \caption{Four Example BARN Environments in the Gazebo Simulator (ordered by ascending relative difficulty level)}.
    \label{fig::barn_worlds}
\end{figure*}

The BARN Challenge started on March 29\textsuperscript{th}, 2022, two months before the ICRA 2022 conference, with a standardized simulation qualifier.
The qualifier used the BARN dataset~\cite{perille2020benchmarking}, which consists of 300 $5\textrm{m}\times5\textrm{m}$ obstacle environments randomly generated by cellular automata (see examples in Fig. \ref{fig::barn_worlds}), each with a predefined start and goal. 
These obstacle environments range from relatively open spaces, where the robot barely needs to turn, to highly dense fields, where the robot needs to squeeze between obstacles with minimal clearance.
The BARN environments are open to the public, and were intended to be used by the participating teams to develop their navigation stack.
Another 50 unseen environments, which are not available to the public, were generated to evaluate the teams' systems.
A random BARN environment generator was also provided to teams so that they could generate their own unseen test environments.\footnote{\url{https://github.com/dperille/jackal-map-creation}}

In addition to the 300 BARN environments, six baseline approaches were also provided for the participants' reference, ranging from classical sampling-based~\cite{fox1997dynamic} and optimization-based navigation systems~\cite{quinlan1993elastic}, to end-to-end machine learning methods~\cite{xubenchmarking, wang2021agile}, and hybrid approaches~\cite{xu2021applr}.
All baselines were implementations of different local planners used in conjunction with Dijkstra's search as the global planner in the ROS \texttt{move\_base} navigation stack~\cite{rosmovebase}.
To facilitate participation, a training pipeline capable of running the standardized Jackal robot in the Gazebo simulator with ROS Melodic (in Ubuntu 18.04), with the option of being containerized in Docker or Singularity containers for fast and standardized setup and evaluation, was also provided.\footnote{\url{https://github.com/Daffan/ros_jackal}}

\subsection{Rules}
Each participating team was required to submit their developed navigation system as a (collection of) launchable ROS node(s).
The challenge utilized a standardized evaluation pipeline\footnote{\url{https://github.com/Daffan/nav-competition-icra2022}} to run each team's navigation system and compute a standardized performance metric that considered navigation success rate (collision or not reaching the goal count as failure), actual traversal time, and environment difficulty (measured by optimal traversal time).
Specially, the score $s$ for navigating each environment $i$ was computed as
\[
s_i = 1^{\textrm{success}}_i \times \frac{\textrm{OT}_i}{\textrm{clip}(\textrm{AT}_i, 4\textrm{OT}_i, 8\textrm{OT}_i)} \; ,
\]
where the indicator function $1_\textrm{success}$ evaluates to $1$ if the robot reaches the navigation goal without any collisions, and evaluates to $0$ otherwise.
$\textrm{AT}$ denotes the actual traversal time, while $\textrm{OT}$ denotes the optimal traversal time, as an indicator of the environment difficulty and measured by the shortest traversal time assuming the robot always travels at its maximum speed ($2\textrm{m/s}$):
\[
\textrm{OT}_i = \frac{\textrm{Path Length}_i}{\textrm{Maximal Speed}}.
\]
The Path Length is provided by the BARN dataset based on Dijkstra's search from the given start to goal.
The $\textrm{clip}$ function clips $\textrm{AT}$ within 4OT and 8OT, in order to assure navigating extremely quickly or slowly in easy or difficult environments respectively won't disproportionally scale the score.
The overall score of each team is the score averaged over all 50 unseen test BARN environments, with 10 trials in each environment. Higher scores indicate better navigation performance.
The six baselines score between $0.1627$ and $0.2334$. The maximum possible score based on our metric is 0.25.

\subsection{Results}
The simulation qualifier started on March 29\textsuperscript{th}, 2022 and lasted through May 22\textsuperscript{th}, 2022.
In total, five teams from all over the world submitted their navigation systems.
The performance of each submission was evaluated by a standard evaluation pipeline, and the results are shown in Tab. \ref{tab::sim_results}. 

\begin{table}[h]
  \caption{Simulation Results}
  \label{tab::sim_results}
  \centering
  \small
  \begin{tabular}{ccc}
  \toprule
  Rank. & Team/Method (University) & Score \\
  \midrule
  1 & TRAIL (Temple University) & 0.2415\\
  2 & LfLH (Baseline~\cite{wang2021agile}) & 0.2334\\
  3 & AMRL (UT Austin) & 0.2310\\
  4 & AMR (UVA) & 0.2200\\
  5 & E-Band (Baseline~\cite{quinlan1993elastic}) & 0.2053\\
  6 & End-to-End (Baseline~\cite{pfeiffer2017perception}) & 0.2042\\
  7 & APPLR-DWA (Baseline~\cite{xu2021applr}) & 0.1979\\
  8 & Yiyuiii (Nanjing University) & 0.1969\\
  9 & NavBot (Indian Institute of Science) & 0.1733\\
  10 & Fast ($2.0$m/s) DWA (Baseline~\cite{fox1997dynamic}) & 0.1709\\
  11 & Default ($0.5$m/s) DWA (Baseline~\cite{fox1997dynamic}) & 0.1627\\
  \bottomrule
  \end{tabular}
\end{table}

All methods outperformed the baseline Dynamic Window Approach (DWA)~\cite{fox1997dynamic}, with both $2.0$m/s and $0.5$m/s max speed, the latter of which is the default local planner for the Jackal robot.
However, only one approach (from Temple University) outperformed all baselines.
The top three teams from the simulation qualifier, i.e., Temple Robotics and Artificial Intelligence Lab (TRAIL) from Temple University, Autonomous Mobile Robotics Laboratory from The University of Texas at Austin (AMRL UT Austin), and Autonomous Mobile Robots Lab from The University of Virginia (AMR UVA), were invited to the physical finals at ICRA 2022.
\section{Physical Finals}
\label{sec::physical}
The physical finals took place at ICRA 2022 in the Philadelphia Convention Center on May 25\textsuperscript{th} and May 26\textsuperscript{th}, 2022.
Two physical Jackal robots with the same sensors and actuators were provided by the competition sponsor, Clearpath Robotics. 

\subsection{Rules}
\begin{figure}
    \centering
    \includegraphics[width=0.8\columnwidth]{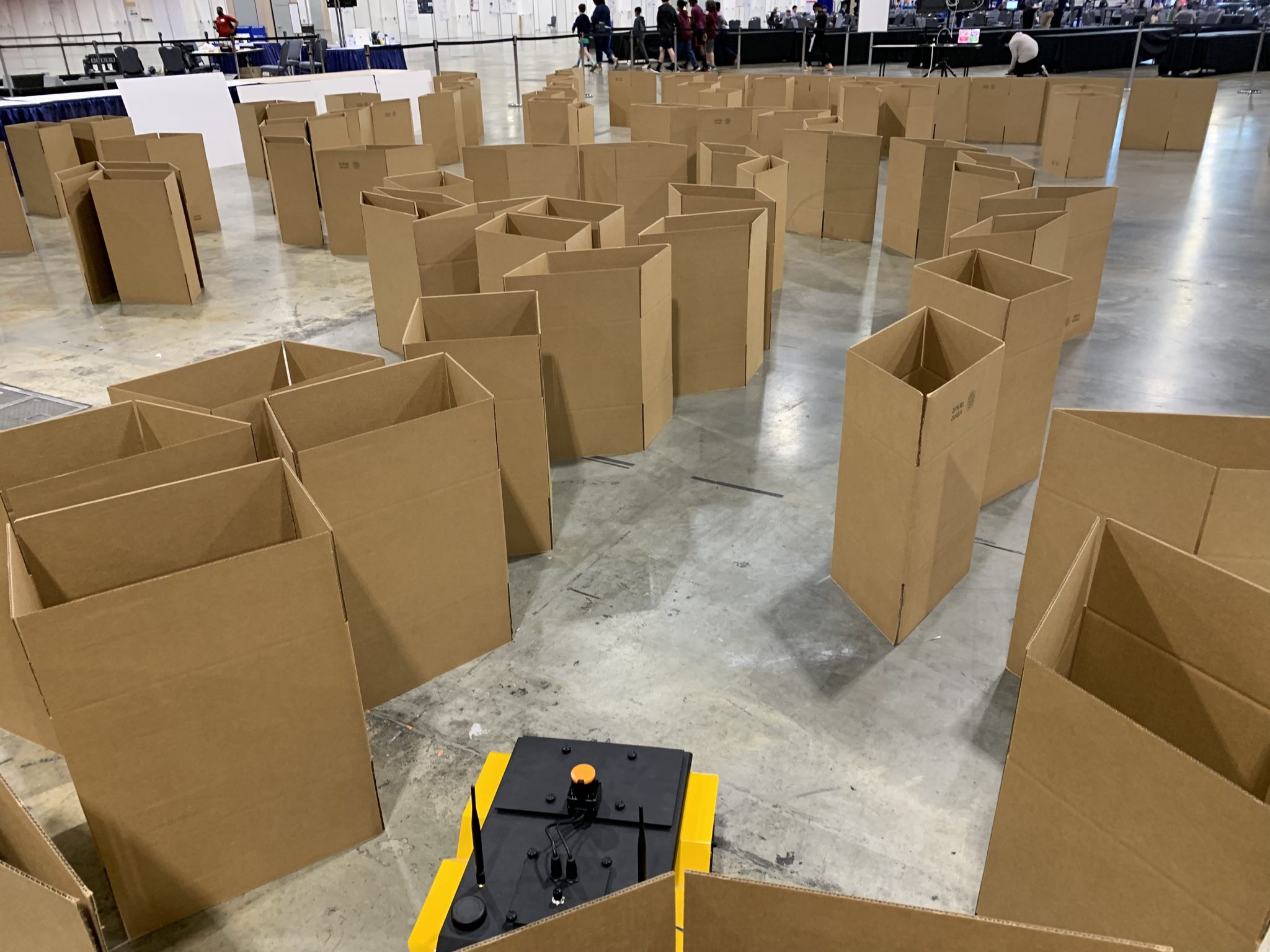}
    \caption{One (Out of Three) Physical Obstacle Courses during the Finals}
    \label{fig::physical_course}
\end{figure}

Physical obstacle courses were set up using approximately 200 cardboard boxes in the convention center (Fig. \ref{fig::physical_course}).
Because the goal of the challenge was to test a navigation system's ability to perform local planning, all three physical obstacle courses have an obvious passage that connects the start and goal locations (i.e., the robot shouldn't be confused by global planing at all), but the overall obstacle clearance when traversing this passage was designed to be very constrained, e.g., a few centimeters around the robot.

While it was the organizers' original intention to run exactly the same navigation systems submitted by the three top teams and use the same scoring metric in the simulation qualifiers in the physical finals, these systems suffered from (surprisingly) poor navigation performance in the real world (not even being able to finish one single trial without any collisions).
Therefore, the organizers decided to change the rules by giving each team 30 minutes before competing in each of the three physical obstacle courses in order to fine-tune their navigation systems.
After all three teams had this chance to set up for a particular obstacle course, the actual physical finals started as a 30-minute timed session for each team.
In each 30-minute session, a team tested their navigation system in the obstacle course and notified the organizers when they were ready to time a competition trial.
Each team had the opportunity to run five timed trials (after notifying the organizers).
The fastest three out of the five timed trials were counted, and the team that had the most successful trials (reaching the goal without any collision) was the winner.
In the case of a tie, the team with the fasted average traversal time would be declared the winner. 

\subsection{Results}
\begin{figure}[h]
    \centering
    \includegraphics[width=1\columnwidth]{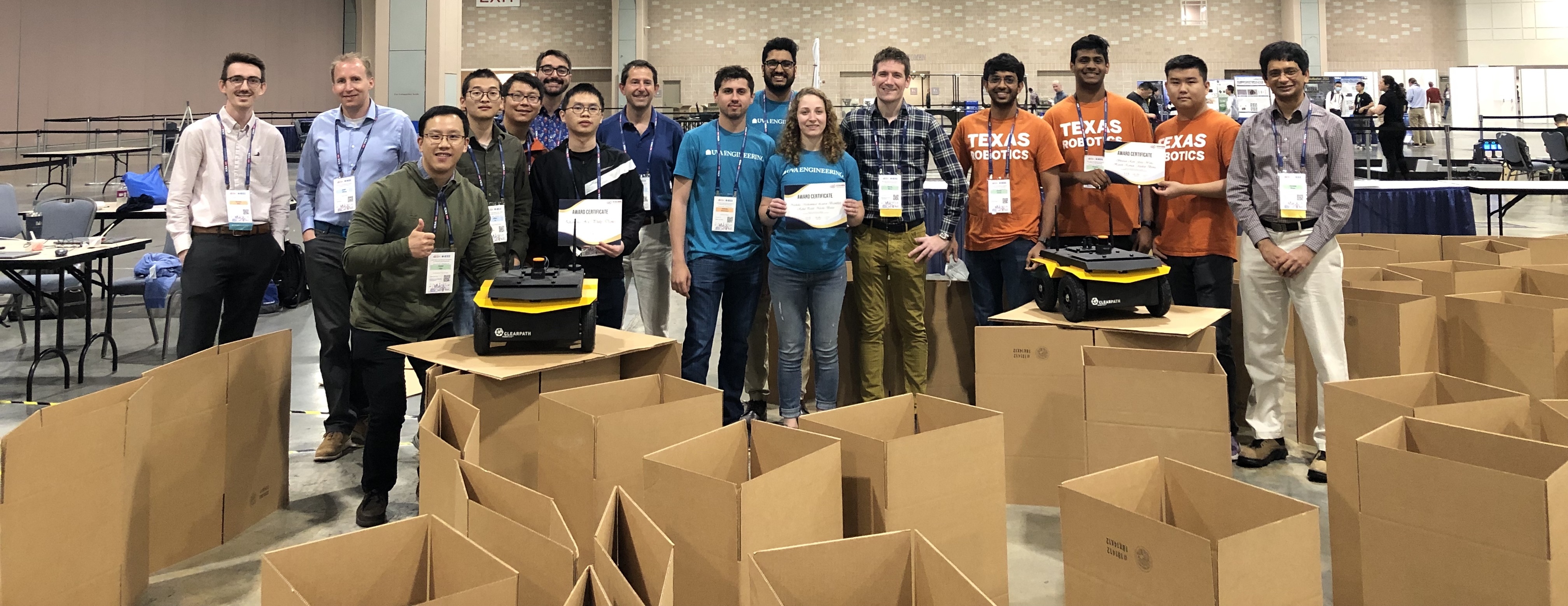}
    \caption{From Left to Right: Competition Sponsor (Clearpath Robotics), Competition Organizers, the Temple, UVA, and UT Austin teams}
    \label{fig::barn_at_icra}
\end{figure}

The physical finals took place on May 25\textsuperscript{th} and May 26\textsuperscript{th}, 2022 (see the final award ceremony in Fig. \ref{fig::barn_at_icra}).
The three teams' navigation performance is shown in Tab. \ref{tab::physical_results}.
Since all navigation systems navigated at roughly the same speed, the final results were determined solely by the success rate of the best three out of five timed trials for each team.
Surprisingly, the best system in simulation by Temple University exhibited the lowest success rate, while UT Austin's system enjoyed the highest rate of success.

\begin{table}[h]
  \caption{Physical Results}
  \label{tab::physical_results}
  \centering
  \small
  \begin{tabular}{ccc}
  \toprule
  Rank. & Team/Method (University) & Success / Total Trials \\
  \midrule
  1 & AMRL (UT Austin) & 8/9\\
  2 & AMR (UVA) & 4/9\\
  3 & TRAIL (Temple University) & 2/9\\
  \bottomrule
  \end{tabular}
\end{table}

\section{Top Three Teams and Approaches}
\label{sec::teams}
In this section, we report the approaches used by the three winning teams.

\subsection{The University of Texas at Austin}
To enable robust, repeatable, and safe navigation in constrained spaces frequently found in BARN, the UT Austin team from AMRL\footnote{\url{https://amrl.cs.utexas.edu/}} utilized state-of-the-art classical approaches to handle localization, planning, and control along with an automated pipeline to visualize and debug continuous integration. To plan feasible paths to reach the goal location while avoiding obstacles, a medium-horizon kinematic planner from \texttt{ROS} \texttt{move\_base} \cite{rosmovebase} was used, combined with a discrete path rollout greedy planner for local kinodynamic planning from AMRL's graph navigation stack \cite{graphnavgithub}. This two-stage hierarchical planning generated safe motion plans for the robot to make progress towards the goal while reactively avoiding obstacles along its path using the LiDAR scans. Additionally, since the environment contains tight spaces that are challenging to navigate through, it was observed that accurate motion estimation of the robot was crucial to deploying a planning-based navigation controller in an unmapped environment. When executing sharp turns in constrained environments, poor estimates of the robot's motion negatively interfered with costmap updates in \texttt{move\_base} and often prevented the mid-level planner from discovering any feasible path to the goal.

Towards addressing this problem, Episodic non-Markov Localization (EnML) \cite{biwasenml} was utilized, which fuses the LiDAR range scans with wheel odometry through non-markov bayesian updates. Combining EnML with two-stage hierarchical planning proved to be useful in safely handling constrained spaces. Additionally, the UT Austin team developed custom automated tools to generate visualizations for debugging that helped identify failure cases easily, perform manual hyperparameter tuning and accelerate bug fixes during the competition. 

While classical approaches helped solve a majority of environments in the BARN challenge, significant challenges still remain for navigation in extremely constrained spaces. For example, the two-stage hierarchical planning module does not explore unobserved regions of the environment before committing to a kinematically feasible path. This sometimes leads to suboptimal paths causing longer time taken to reach the goal. We posit that a learnable mid-level planner with the ability to actively explore the environment appropriately to plan the optimal path may be a promising future direction of research to improve autonomous navigation in constrained spaces.

\subsection{University of Virginia}
\begin{figure}[ht!]
\centering
\subfigure[]{\label{fig::gap_ex}\includegraphics[width=.23\textwidth]{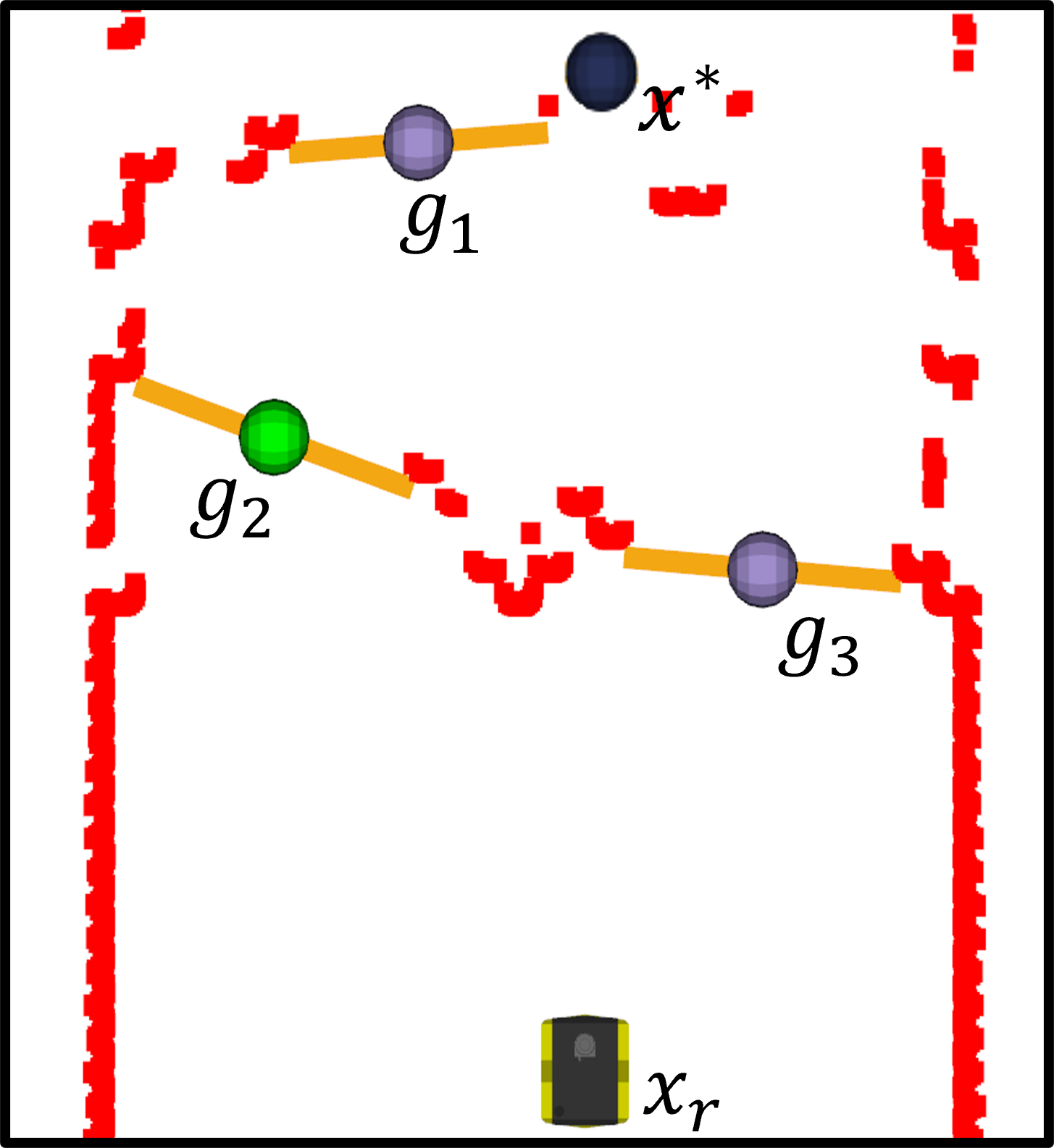}}
\subfigure[]{\label{fig::local_plan}\includegraphics[width=.23\textwidth]{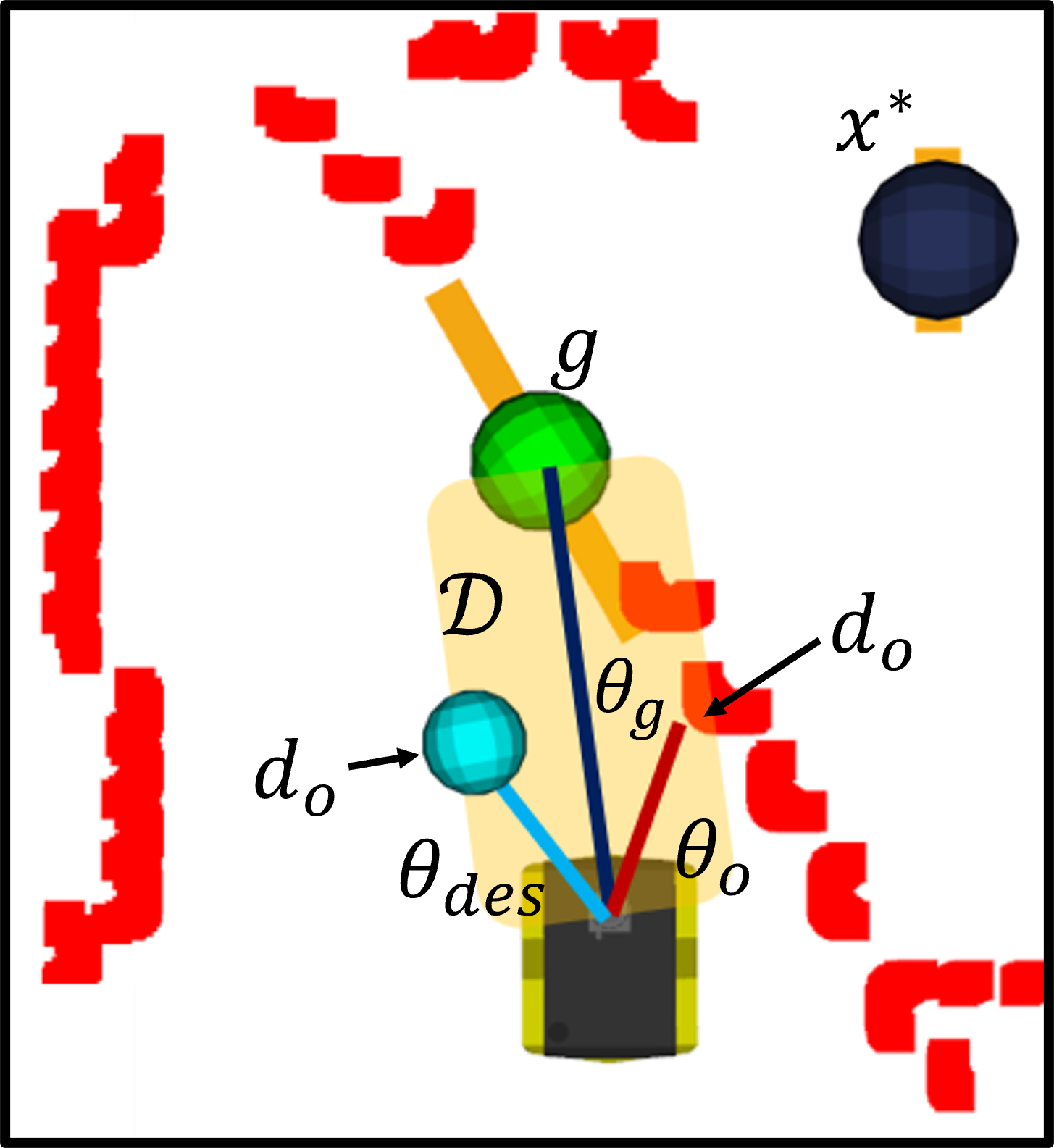}}
\caption{(UVA Team) Examples of (a) Detected Gaps in a Simulated BARN Environment and (b) Local Planner Obstacle Avoidance}
\end{figure}

In order to quickly and robustly navigate through the unknown, cluttered BARN challenge environments, the UVA AMR team\footnote{\url{https://www.bezzorobotics.com/}} developed a mapless, ``follow-the-gap" planning scheme which (a) detects open gaps for the robot to follow to reach a final goal and (b) plans local goals in order to reach these open gaps without colliding with intermediate obstacles. The framework expands upon the UVA AMR lab's previous work \cite{mohammad2022occlusion}. Fig.~\ref{fig::gap_ex} illustrates the framework displaying the laser scan point-cloud of a world from the BARN dataset along with the detected intermediate gaps $g_1,\ g_2,$ and $g_3$, vehicle position $x_r \in \mathbb{R}^2$, and final goal position $x^* \in \mathbb{R}^2$. Fig. ~\ref{fig::local_plan} shows the local planner, which provides course corrections in order for the robot to avoid obstacles while reaching a selected gap goal.
The approach takes advantage of the fact that gaps start or end at discontinuities in the laser scan and leverages this principle to find intermediate gap goals for navigation \cite{mujahad2010gaps}. Let $p_i, p_{i+1} \in \mathbb{R}^2$ be adjacent points in the laser scan and $R$ denote the maximum sensing range of the LiDAR. The first discontinuity, referred to as \textit{Type 1}, occurs when the distance between the adjacent readings is larger than the circumscribed diameter $d_r$ of the robot: $||p_i-p_{i+1}||_2 > d_r$. The second discontinuity, \textit{Type 2}, occurs when one of the two readings is outside the LiDAR's sensing range: $||p_i-x_r||_2 \geq R \oplus ||p_{i+1}-x_r||_2 \geq R$ . If $||p_{i+1}-x_r||_2 > ||p_i-x_r||_2$, the discontinuity is referred to as \textit{rising}, otherwise it is \textit{falling}. Below we describe how to leverage these discontinuities to identify gaps.

The first step in gap detection is to perform a forward pass from $p_0$ to $p_{n-1}$ in the laser point-cloud scan for rising type 1 and type 2 discontinuities. Let $p_i$ denote the location of the first rising discontinuity and $\mathcal{L}^+=\{i+1,\dots,n-1\}$. This point becomes the beginning of the gap. To determine the end, we find the point $p_j$ closest to $p_i$ such that $j \in \mathcal{L}^+$. That is,
\begin{equation}
    p_j = \arg\min_{j \in \mathcal{L}^+} ||p_i - p_j||_2
\end{equation}

The process continues starting from $p_{j+1}$. Once the forward pass is complete, a mirrored backward pass from $p_{n-1}$ to $p_0$ is done to find gaps via falling discontinuities. Each detected gap, defined as $g_i=(a_i,b_i)$, a tuple of the start and end points, are added to a quadtree $\mathcal{T}_g$ which keeps track of where all previously identified and visited gaps are located. If any gap already exists in the tree, it is ignored.

Once $\mathcal{T}_g$ is updated, a gap $g^* \in \mathcal{T}_g$ is selected to be the intermediate goal if it is determined that the final goal $x^*$ is not \textit{admissible}. In this context, admissibility is determined by checking if a given goal is navigable; that is, from the laser scan data, a path is known to exist from the robot position to the goal. The check is done by using a similar algorithm as discussed by Minguez and Montano \cite{minguez2004ND}, which, given any start point $x_a$ and end point $x_b$, ensures no inflated obstacles block the robot along the line connecting the two points.

The process to select the gap goal from $\mathcal{T}_g$ when $x^*$ is inadmissible is outlined in Algorithm \ref{alg::FindGapGoal}. At each iteration, the algorithm finds the closest gap $g^*$ to the final goal $x^*$. If $g^*$ is inadmissible from the robot's current position, properties of quadtree queries are utilized to find all gaps $G' \subseteq \mathcal{T}_g$ which must be passed as the robot drives from $x_r$ to $g^*$. The algorithm then iteratively finds the closest admissible gap $g \in G'$ to the robot which is also admissible to $g^*$. Meaning, the robot knows that a feasible path from $x_r$ to $g$ and from $g$ to $g^*$ exists. If no $g$ satisfy this constraint for the given $g^*$, the process repeats with $g^*$ as the next closest gap to $x^*$ and terminates once an admissible gap is found. For clarity, Fig. \ref{fig::gap_ex} shows an example of the goal selection process. The final goal $x^*$ is not admissible, nor is the closest gap to it, $g_1$. However, $x_r$ to $g_2$ is admissible as well as $g_2$ to $g_1$. Thus, $g_2$ is selected as the intermediate goal and the selection process repeats once the robot reaches $g_2$.

\begin{algorithm}[ht!]
\caption{(UVA Team) Find Gap Goal}
\label{alg::FindGapGoal}
\begin{algorithmic}[1]
\STATE \textbf{Input:} quadtree $\mathcal{T}_g$, robot position $x_r$, final goal $x^*$
\STATE \textbf{Output:} gap goal $g^*$
\WHILE {$\mathcal{T}_g \not= \emptyset$ \& $\text{!isAdmissible}(x_r,g^*)$ }
\STATE $g^* \leftarrow \arg\min_{g^* \in \mathcal{T}_g} ||x^*-g^*||_2$
\STATE $\mathcal{T}_g \leftarrow \mathcal{T}_g \setminus \{g^*\}$
\STATE \textit{\# Returns children in descending order of dist. to $x_r$}
\STATE $G' \leftarrow \text{getChildren}(g^*,x_r,\mathcal{T}_g)$
\FOR{$g \in G'$}
\IF{$\text{isAdmissible}(x_r,g)$ \& $\text{isAdmissible}(g,g^*)$}
\STATE $g^*=g$
\ENDIF
\ENDFOR
\ENDWHILE
\RETURN $g^*$
\end{algorithmic}
\end{algorithm}
Even though the selected gap goal is admissible, a direct path to it may not be feasible given the configuration of the obstacles within an environment. For example, a robot navigating directly to $g$ in Fig.~\ref{fig::local_plan} will collide with the obstacles shown by the laser scan data. In order to prevent such issues from arising, local planner is utilized which re-plans the mobile robot's trajectory at every timestep if collision is imminent. The direct path to the goal is formulated as a region $\mathcal{D}$, which accounts for the relative heading to the goal, $\theta_{g}$ and the diameter $d_r$ of the robot. The region $\mathcal{D}$ is checked against the laser scan points for any obstacles; let $\bm{p}$ represent all obstacle coordinates within region $\mathcal{D}$. If no obstacles are in $\mathcal{D}$, that is $\bm{p} = \emptyset$, the robot is sent directly to the gap goal, $g$. If there are multiple obstacles within $\mathcal{D}$, the one closest to the robot is selected; let $d_o$ represent the distance to the closest obstacle and $\theta_o$ represent the direction of the obstacle with respect to the robot's heading. The new desired heading is then computed by accounting the offset between goal and obstacle to the gap goal: $\theta_{des} = \theta_g + (\theta_g-\theta_o)$, and the local goal is placed at a distance of $d_o$ in this desired direction (shown in teal in Fig.~\ref{fig::local_plan}).

The inputs to the robot are angular and linear velocities, and are determined using proportional controllers:
\begin{equation}
    \begin{cases} 
      \omega = \min(k_t(\theta_{des}-\theta_r),\omega_{\max}), \\
      v = k_vv_{\max}(1-\alpha\frac{|\omega|}{\omega_{\max}})
   \end{cases}
\end{equation}
where $k_t$, $k_v$, and $\alpha$ are constant proportional gains, $\theta_r$ is the current heading of the robot, and $\omega_{\max}$ and $v_{\max}$ are the maximum angular and linear velocities respectively.

\subsection{Temple University}
\begin{figure*}[t]
    \centering
    \includegraphics[width=0.8 \textwidth]{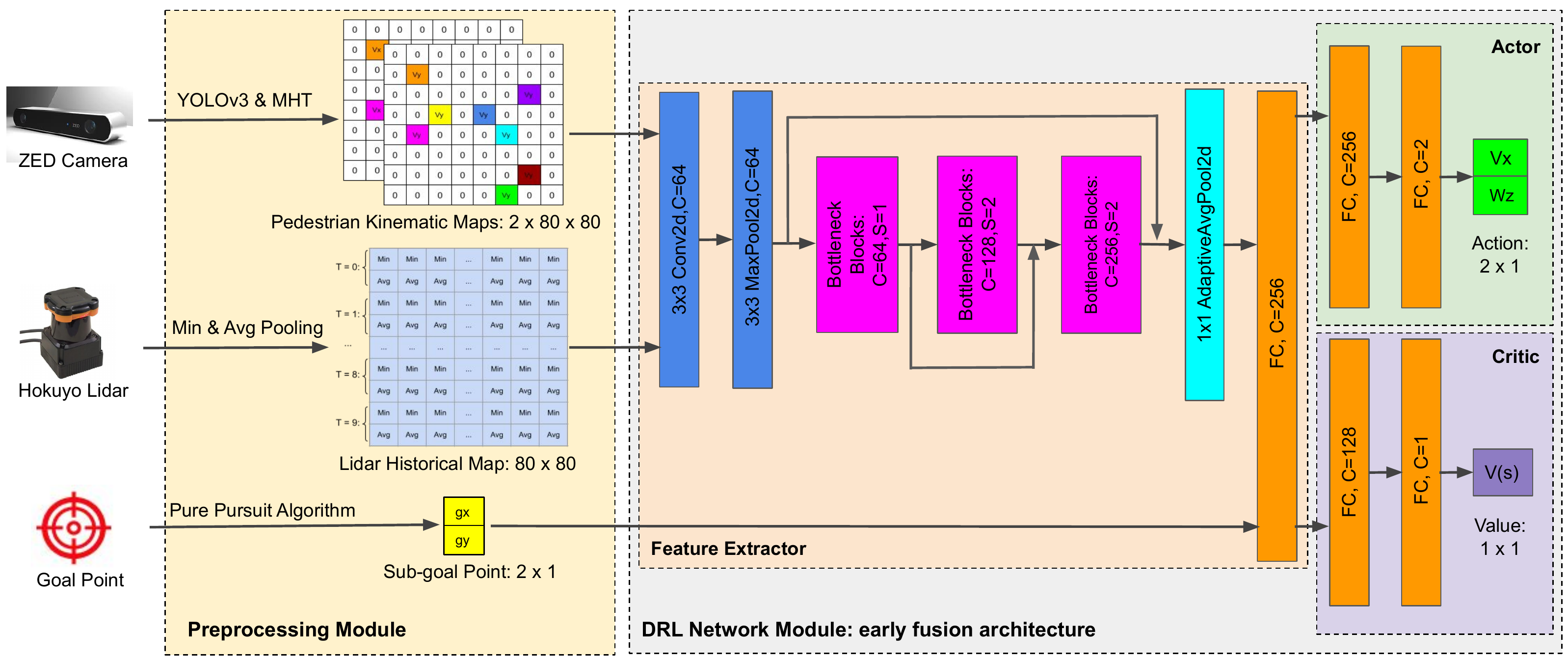}
    \caption{(Temple Team) The system architecture of the DRL-VO control policy. 
    Raw sensor data from the ZED camera and Hokuyo LiDAR, as well as the goal point, are fed into a preprocessing module to create intermediate data representations.
    These low-level intermediate features are fused in a feature extractor network to obtain high-level abstract features.
    The actor network uses these abstract features to generate steering commands to control the robot, while the critic network outputs the state value for training the policy. 
    }
    \label{fig:drl_vo}
\end{figure*}

The team at Temple\footnote{\url{https://sites.temple.edu/trail/}} used a deep reinforcement learning (DRL) based control policy, called DRL-VO \cite{xie2022drl}, originally designed for safe and efficient navigation through crowded dynamic environments.
The system architecture of the DRL-VO control policy, shown in Fig.~\ref{fig:drl_vo}, is divided into two modules: preprocessing and DRL network.

\subsubsection{Preprocessing Module}
Instead of directly feeding the raw sensor data into deep neural networks like other works \cite{pfeiffer2017perception,long2018towards,fan2020distributed, guldenring2020learning}, the DRL-VO control policy utilizes preprocessed data as the network input.
There are three types of inputs that capture different aspects of the scene.
\begin{enumerate}
    \item \textbf{Pedestrians}: To track pedestrians, the raw RGB image data and point cloud data from a ZED camera are fed into the YOLOv3 \cite{redmon2018yolov3} object detector to get pedestrian detections. These detections are passed into a multiple hypothesis tracker (MHT) \cite{yoon2018multiple} to estimate the number of pedestrians and their kinematics (i.e., position and velocity). These pedestrian kinematics are encoded into two $80 \times 80$ occupancy grid-style maps.
    
    \item \textbf{Scene Geometry}: To track the geometry, the past 10 scans (\unit[0.5]{s}) of LiDAR data are collected. Each LiDAR scan is downsampled using a combination of minimum pooling and average pooling, and these downsampled LiDAR data are then reshaped and stacked to create an $80 \times 80$ LiDAR map.

    \item \textbf{Goal Location}: To inform the robot where to go, the final goal point and its corresponding nominal path are fed into the pure pursuit algorithm \cite{coulter1992implementation} to extract the sub-goal point, which is fed into the DRL-VO network.
\end{enumerate}
This novel preprocessed data representation is one key idea of the DRL-VO control policy, allowing it to bridge the sim-to-real gap and generalize to new scenarios better than other end-to-end policies.

\subsubsection{DRL Network Module}
The DRL-VO control policy uses an early fusion network architecture to combine the pedestrian and LiDAR data at the input layer in order to obtain high-level abstract feature maps.
This early fusion architecture facilitates the design of small networks with fewer parameters than late fusion works \cite{sathyamoorthy2020densecavoid, huang2021towards}, which is the key deploying them on resource-constrained robots.
These high-level feature maps are combined with the sub-goal point and fed into the actor and critic networks to generate suitable control inputs and a state value, respectively.

\subsubsection{Training} 
To ensure that the DRL-VO policy leads the robot to navigate safely and efficiently, the team at Temple developed a new multi-objective reward function that rewards travel towards the goal, avoiding collisions with static objects, having a smooth path, and actively avoiding future collisions with pedestrians.
This final term, which utilizes the concept of velocity obstacles (VO) \cite{fiorini1998motion, wilkie2009generalized}, is key to the success of the DRL-VO control policy.
With this reward function, the DRL-VO policy is trained via the proximal policy optimization (PPO) algorithm  \cite{schulman2017proximal} in a 3D lobby Gazebo simulation environment with 34 moving pedestrians using a Turtlebot2 robot with a ZED camera and a 2D Hokuyo LiDAR.

\subsubsection{Deployment}
The Temple team directly deployed the DRL-VO policy trained on a Turtlebot2 in The BARN Challenge without any model fine-tuning. To achieve this, the team had to account for three key differences:
\begin{enumerate}
    \item \textbf{Unknown Map}: During development, the DRL-VO policy used a known occupancy grid map of the static environment for localization, which is not available in the BARN challenge. To account for this, the localization module (\texttt{amcl}) was removed from the software stack and replaced with a laser odometry module.
    
    \item \textbf{Static Environment}: The DRL-VO policy was designed to function in dynamic environments. To account for the fact that the environments in the BARN Challenge were all static and highly constrained, the pedestrian map was set to all zeros.
    
    \item \textbf{Different Robot Model}: The DRL-VO policy was trained on a Turtlebot2, which has a different maximum speed and footprint compared to the Jackal platform. In the BARN Challenge, the maximum speed of the robot was modified based on its proximity to obstacles. This led to the robot moving slowly (\unit[0.5]{m/s}, same speed as the Turtlebot2) when near obstacles and quickly (up to \unit[2]{m/s}, maximum speed of the Jackal) when in an open area.
\end{enumerate}
\section{Discussions}
\label{sec::discussions}
Based on each team's approach and the navigation performance observed during the competition, we now discuss lessons learned from The BARN Challenge and point out promising future research directions to push the boundaries of efficient mobile robot navigation in highly constrained spaces. 

\subsection{Generalizability of Learning Based Systems}
One surprising discrepancy between the simulation qualifier and the physical finals is the contrasting performance of the DRL-VO approach by Temple University, which outperformed all baselines and other participants in simulation, but suffers from frequent collision with obstacles in the real world. Despite the fact that the organizers modified the rules during the physical finals to allow the teams to make last-minute modifications to their navigation systems, DRL-VO still did not perform well in all three physical obstacle courses. The TRAIL team believes this is due to two types of {\em gap} between the simulator and the real world: 1) the real world environments were all highly constrained, whereas the simulator environments contained both constrained and unconstrained environments, and 2) the DRL-VO policy was learned on a Turtlebot2 model (which has a smaller physical footprint than a Jackal). Most of the collisions during the hardware tests were light grazes on the side, so a robot with a smaller size may have remained collision-free.

The stark performance contrast between simulation and the real world suggests a generalizability gap for the reinforcement learning approach. It was not practical for the team to re-train a new system on site during the competition, given the impractically massive amount of training data required for reinforcement learning. How to address this generalizability gap and to make a navigation policy trained in simulation generalizable to the real world and different robot/sensor configurations remains to be investigated, even for such a simple, static obstacle avoidance problem.

Another potential way to address such inevitable generalizability gaps is to seek help from a secondary classical planner that identifies out-of-distribution scenarios in the real world and recovers from them through rule-based heuristics. In fact, for the last two physical courses, the Temple team tried to implement just such a ``recovery planner'' as a backup for DRL-VO: when a potential collision is detected (i.e., the robot faces an obstacle that is too close), the robot rotates in place to head towards an empty space in an attempt to better match the real-world distribution to that in the simulation during training. Although such a recovery planner did help in some scenarios, it is difficult for it to cover every difficult scenario and navigate through the entire obstacle course. Indeed, the Temple team spent time during the 30-minute timed sessions to fine tune the parameters of the recovery planner, but found it difficult to find a single set of parameters to recover the robot from all out-of-distribution scenarios while not to accidentally drive the robot into such scenarios throughout the entire course. On one hand, the simple nature of the recovery planner designed onsite during the competition contributed to the failure. On the other hand, tuning parameters of a planner to cover as many scenarios as possible remains a difficult problem, and will be discussed further below.

\subsection{Tunability of Classical Systems}
Similar to Temple's rule-based recovery planner, UT Austin team's entire navigation system relies on classical methods: EnML localization, medium-horizon kinematic planner, and local rollout-based kinodynamic planner. Inevitably, these classical approaches have numerous tuning parameters, which need to be correctly tuned to cover as many scenarios as possible. A natural disadvantage of relying on a single set of parameters to cover all different difficult scenarios in the BARN Challenge (e.g., dense obstacle fields, narrow curving hallways, relatively open spaces) is the inevitable tradeoff or compromise to sacrifice performance in some scenarios in order to succeed in others or to decrease speed for better safety. Indeed, the UT Austin team's strategy in the physical finals is to spend the first 20 minutes in the 30-minute timed session to fine tune the system parameters until a good set of parameters that allow successful navigation through the entire obstacle course is found, then finish three successful ``safety trials'' first, and finally re-tune the system to enable faster, more aggressive, but riskier navigation behaviors to reduce average traversal time. Although most such ``speed trials'' failed, luckily for the UT Austin team, other teams' inability to safely finish three collision-free trials to the goal make them the winner of the BARN Challenge only with a higher success rate (not faster navigation). 

Two orthogonal future research directions can potentially help with the tunability of navigation systems: (1) developing planners free of tunable parameters onsite during deployment, such as end-to-end learning approaches, but, as mentioned above, with significantly better sim-to-real transfer and generalizability; (2) enabling more intelligent parameter tuning of classical systems, rather than laborious manual tuning, for example, through automatic tuning \cite{ma2021navtuner} or even dynamic parameter policies \cite{xiao2022appl} learned from teleoperated demonstration \cite{xiao2020appld}, corrective interventions \cite{wang2021appli}, evaluative feedback \cite{wang2021apple}, or reinforcement learning \cite{xu2021applr}. 

\subsection{Getting ``Unstuck''}
Although most of the failure trials during the physical finals were due to collision with obstacles, there were also many trials that did not succeed because the robot got stuck in some densely populated obstacle areas. In those places, the robot kept repeating the same behaviors multiple times, e.g., detecting imminent collision with obstacles, rotating in place, backing up, resuming navigation, detecting the same imminent collision again, and so on. Such behavior sometimes led to collision with an obstacle, sometimes got the robot stuck forever, and may also succeed in rare occasions. All three teams have experienced such behaviors, with the UT Austin and UVA teams being able to fix it by tuning parameters and the Temple team changing the threshold between DRL-VO and the recovery planner. 

Similarly, in real-world autonomous robot navigation, how to get ``unstuck'' safely remains a common and challenging problem. No matter how intelligent an autonomous mobile robot is, it may still make mistakes in the real world, e.g., when facing scenarios out of the training distribution, corner cases not considered by the system developer, or situations where the current parameter set is not appropriate. It is very likely that the robot will repeat the same mistake over and over, e.g., getting stuck at the same place, which needs to be avoided. Future research should investigate ways to identify such ``stuck'' situations, balance the tradeoff between exploitation and exploration (i.e., when to keep trying the previous way vs. when to try out new ways to get unstuck), utilize previous successful exploratory experiences in future similar scenarios to not get stuck again \cite{liu2021lifelong}, or leverage offline computation to correct such failure cases in the future~\cite{xulearning}.

\subsection{Tradeoff between Safety and Speed}
While The BARN Challenge was originally designed to test existing navigation system's speed of maneuvering through highly constrained obstacle environments, given the safety constraint of being collision-free, it ended up being a competition about safety alone. The UT Austin team won the competition simply by safely navigating eight out of nine physical trials, not by doing so with the fastest speed. All the teams, except the UT Austin team after they figured out an effective set of parameters for each physical obstacle course, struggled with simply reaching the goal without any collision. The challenge organizers also deployed the widely-used DWA planner \cite{fox1997dynamic} in the ROS \texttt{move\_base} navigation stack in the physical obstacle courses, only to find out that, despite being relatively safe compared to the participating teams' methods, it struggled with many narrow spaces and got stuck in those places very often. Such a fact shows that the current autonomous mobile robot navigation capability still lags farther behind than one may expect.

\subsection{Latency Compensation for High Speed}
Only the UT Austin team attempted to pursue higher speed navigation ($>0.5$ m/s), doing so after an appropriate parameter set was found for the particular physical course and three successful ``safety trials'' have been achieved. However, most ``speed trials'' ended in collision. One contributing factor to such failure was improper latency compensation for various high speeds. The UT Austin team was the only team that explicitly considered latency compensation in their AMRL stack \cite{graphnavgithub}, through a latency parameter. During high-speed maneuvers, the robot inevitably needs to aggressively change its navigation speed to swerve through obstacles and to accelerate in open spaces. System latency caused by sensing, processing, computation, communication, and actuation will likely invalidate previously feasible plans. While simply tuning the latency parameter value can help to certain extent, a more intelligent and adaptive way to calculate and compensate system latency is necessary for the robot to take full advantage of its computing power before executing aggressive maneuvers.

\subsection{Navigation is More Than Planning}
To plan agile navigation maneuvers through highly constrained obstacle environments, the robot first needs to accurately perceive its configuration with respect to the obstacles. 
Inaccurate localization or odometry during fast maneuvers with significant angular velocity usually produces significant drift, causing previously valid plans become infeasible. While all three teams' local planners rely on raw perception to minimize such adverse effect, e.g., using high frequency laser scans and directly planning with respect to these raw features, their global planner usually depends on the results of localization or odometry techniques. For example, the Temple team used the Dijkstra's global planner in \texttt{move\_base}. An erroneous localization will cause an erroneous global plan, which in turn will affect the quality of the local plan. Such adverse effect will diminish when the navigation speed is low, because localization techniques may recover from drift over time. During high-speed navigation, however, the planner needs to quickly plan actions regardless of whether the drift has been fixed or not. As mentioned above, latency will start to play a role as well, because a good latency compensation technique will depend on an accurate localization and odometry model of the robot, i.e., being able to predict where the robot will be based on where the robot is and what action will be executed. Techniques for better odometry, localization \cite{biwasenml}, and kinodynamic models \cite{xiao2021learning, karnan2022vi, atreya2022high} during high-speed navigation will be necessary to allow mobile robots to move both fast and accurately at the same time. 
\section{CONCLUSIONS}
\label{sec::conclusion}
The results of The BARN Challenge at ICRA 2022 suggest that, contrary to the perception of many in the field, autonomous metric ground robot navigation can not yet be considered a solved problem.
Indeed, even the competition organizers had initially assumed that obstacle avoidance alone was too simple a goal, and therefore emphasized navigation speed before the physical competition.
However, each of the finalist teams experienced difficulty performing collision-free navigation, and this ultimately led the organizers to modify the competition rules to focus more on collision avoidance.
This result suggests that state-of-the-art navigation systems still suffer from suboptimal performance due to potentially many aspects of the full navigation system (discussed in Section \ref{sec::discussions}).
Therefore, while it is worthwhile to extend navigation research in directions orthogonal to metric navigation (e.g., purely vision-based, off-road, and social navigation), the community should also not overlook the problems that still remain in this space, especially when robots are expected to be extensively and reliably deployed in the real world.

\bibliographystyle{IEEEtran}
\bibliography{IEEEabrv,references,references-Temple}
\end{document}